\documentclass{IEEEtaes}
\pdfoutput=1
\usepackage{color,array,amsthm,amsmath,booktabs,multirow,tikz,subcaption, amssymb, amsmath,threeparttable}
\usepackage{pgfplots}
\pgfplotsset{compat=1.18}
\usepackage{xcolor}
\usepackage{subcaption}
\usepackage{graphicx}

\jvol{XX}
\jnum{XX}
\jmonth{XXXXX}
\paper{1234567}
\pubyear{2024}
\doiinfo{TAES.2022.Doi Number}

\begin{document}

\title{Aeroengine performance prediction using a physical-embedded data-driven method}

\author{Tong Mo}
\affil{Institute of Automation, Chinese Academy of Sciences, Beijing, China} 

\author{Shiran Dai}
\author{An Fu}
\affil{Beijing Aerospace Technology Research Institute, Beijing, China} 

\author{Xiaomeng Zhu}
\author{Shuxiao Li}
\affil{Institute of Automation, Chinese Academy of Sciences, Beijing, China}


\receiveddate{Manuscript received XXXXX 00, 0000; revised XXXXX 00, 0000; accepted XXXXX 00, 0000.\\
This work was supported in part by the National Natural Science Foundation of China under Grant 62076020.}

\authoraddress{Authors’ addresses: Tong Mo is with Institute of Automation, Chinese Academy of Sciences (CASIA), Beijing 100190, China, and also with Beijing University of Technology (BJUT), Beijing 100124, China, E-mails: motong02@outlook.com; Shiran Dai and An Fu are with Beijing Aerospace Technology Research Institute, Beijing 100191, China, E-mails: dsr0627clara@163.com, 2982561014@qq.com ; Xiaomeng Zhu and Shuxiao Li are with the School of Artificial Intelligence, University of Chinese Academy of Sciences (UCAS), Beijing 100049, China, and also with the State Key Laboratory of Multimodal Artificial Intelligence Systems (MAIS), Institute of Automation, Chinese Academy of Sciences (CASIA), Beijing 100190, China, E-mails: \{zhuxiaomeng2021, shuxiao.li\}@ia.ac.cn. (Corresponding authors: Shiran Dai and Shuxiao Li.)}

\editor{}

\markboth{AUTHOR ET AL.}{SHORT ARTICLE TITLE}
\maketitle

\begin{abstract}
Accurate and efficient prediction of aeroengine performance is of paramount importance for engine design, maintenance, and optimization endeavours. However, existing methodologies often struggle to strike an optimal balance among predictive accuracy, computational efficiency, modelling complexity, and data dependency. To address these challenges, we propose a strategy that synergistically combines domain knowledge from both the aeroengine and neural network realms to enable real-time prediction of engine performance parameters. Leveraging aeroengine domain knowledge, we judiciously design the network structure and regulate the internal information flow. Concurrently, drawing upon neural network domain expertise, we devise four distinct feature fusion methods and introduce an innovative loss function formulation. To rigorously evaluate the effectiveness and robustness of our proposed strategy, we conduct comprehensive validation across two distinct datasets. The empirical results demonstrate :(1) the evident advantages of our tailored loss function; (2) our model's ability to maintain equal or superior performance with a reduced parameter count; (3) our model's reduced data dependency compared to generalized neural network architectures; (4)Our model is more interpretable than traditional black box machine learning methods.
\end{abstract}


\section{INTRODUCTION}
Aeroengine performance prediction, especially engine thrust simulation, is a critical tool spanning the lifecycle from engine design, integration and control development to real-time performance monitoring and diagnostics of operational engines. It allows manufacturers to optimize designs, ground crews to understand engine health, engine health management, and operators to maximize performance, safety and efficiency. Given the importance of prediction as described above and the development of modern computer science, many methods have emerged.
\par
In general, the simulation based on Computational Fluid Dynamics (CFD) is the most popular way to get pneumatic engine simulation data. Firstly, the CFD model of a specific engine is established by CAD-Computer Aided Design modelling of the specific engine and the creation of a high-quality computational mesh of the fluid volume. Then the CFD simulation is initialized according to the above settings. Finally, the specified parameters are solved by physical equations based on the obtained CFD simulation data \cite{10.1115/1.4049410,en13112846,osti_1765177}. With advances in computing power, many improvements to the CFD simulation have been given \cite{laskowski2016future,aerospace10090782,ma_2019,JIA2022107429} to further increase the accuracy of parameter prediction. These methods usually focus on optimizing the progress of CFD simulation such as developing more advanced turbulence modelling to improve the accuracy of separation flow and heat transfer predictions, thus the accuracy of parameter predictions is also improved. On the whole, CFD-based methods can provide highly accurate simulation data. However, it has a fatal flaw of being very processor-intensive due to the complexity of the underlying mathematics. This leads to substantial computational time and modelling costs, especially for large and detailed models. Although there have been many attempts to optimize for this, such as the use of massively parallel computation, the use of improved numerical algorithms \cite{shang2019landmarks,luo2024immersed,RIZZI2021106940} or the tighter integration of CFD into the design\cite{doi:10.2514/1.A35807}, none of these have changed the fact that CFD is computationally and temporally very costly.
\par
With the development of artificial intelligence, data-driven machine learning(ML) methods have been used to predict engine parameters. ML methods discard physical equations, real engine structures, etc., and treat the whole simulation process as a black box, focusing only on the input and output results. As long as enough high-quality training data can be provided, ML models can learn nonlinear relationships between inputs and outputs within the model independently \cite{aerospace9020060}. As a result, ML methods significantly lower the threshold for accessing simulation data, while CFD methods require deep domain knowledge. In addition, ML methods drastically reduce the time and computational resource cost of obtaining simulation data, which can be orders of magnitude faster than high-fidelity CFD simulations once trained. Specifically, CFD simulations often require many hours on large compute clusters, while ML methods can make real-time predictions \cite{LECLAINCHE2023108354}.
\par
The most typical ML method is the artificial neural networks (ANNs). ANNs have been attempted to be applied to system control and engine diagnosis since as early as the 1970s \cite{sellers1975dyngen,veres2002overview}. Dalkiran \cite{yildirim2021predicting} used ANNs to predict the thrust of an aeroengine using only three inputs: latitude, airspeed, and temperature, and controlled the relative error to less than 5\%. Some comprehensive surveys give a detailed description of the application and development of ANNs for industrial engines as well \cite{FAST20099, nikpey2014experimental}. Li et al.\cite{li2017modelling} compared the performance of three ANNs in predicting the dust removal efficiency of rotating packed beds admitting the modelling difficulties of CFD methods in some cases and the great potential shown by ANNs. Kim et al. \cite{KIM2020117046} verified that ANNs show high accuracy in the transient performance analysis and further improved the data-driven model based on ANNs for transient simulation. Besides ANNs, decision trees, random forests, support vector machines, and clustering (e.g., k-means, hierarchical) are also used to make predictions\cite{aerospace10010017}.  However, the performance of these methods, which are entirely data-driven and discard physical constraints, is highly dependent on the quality and quantity of the training data, resulting in lower model stability compared to that of CFD methods.
\par
To improve the model stability of ML methods, hybrid methods which mix physical properties in a data-driven approach are introduced. There are three main ways to add physical features to a data-driven model\cite{karniadakis2021physics}. The first is focusing on preprocessing the training data, which deepens the physical principles in the data by processing the training data with data augmentation\cite{van2001art}. The second starts from the loss function to enhance the physical features of the model by establishing a connection between the physical formulas and engine principles with the model loss function or injecting extra physical information to influence the model during the period of training\cite{raissi2019physics,robinson2022physics,liu2022physics}. For example, J. Raymond et al. \cite{raymond2021applying} designed a novel loss function that introduces physical formulas for specific cases and datasets. The last and most commonly used one is to make a special design of neural network structure based on domain knowledge\cite{bronstein2017geometric,cuomo2022scientific}. For example, Lin. \cite{lin2023thrust} proposed a hybrid method by fusing the structure of a real aeroengine and nerual network to predict the thrust. Compared to conventional neural networks, the hybrid network has maximum relative deviations below 4.0\% and average relative deviations below 2.0\% on all testing datasets which is superior to conventional neural networks. Additionally, physics-guided neural networks \cite{karpatne2017physics} use the simulated output and observed features based on the physical model to generate the prediction results through the neural network architecture. Despite achieving success, the above methods are either not sufficiently influential in terms of methodology, or cannot be realized in the short term due to the complexity of the physical equations and principles, or the overall design is unreasonable and ineffective due to the inability to balance the domain knowledge in the field of aeroengine and machine learning.

\par
In this research, we will integrate neural network domain knowledge and aeroengine domain knowledge to design a physical-embedded neural network framework in a concise and universal way, so that it is very easy to extend for different applications. As the core of the proposed framework, four distinct feature fusion modules are deliberately designed for effective information aggregation on the premise of minimizing the number of network parameters as much as possible. We also proposed a novel loss function, termed as mean absolute relative error, to balance the optimization effects for both large-valued samples and small-valued samples, thus improving the effectiveness of network training. Extensive experiments have verified that the proposed hybrid model as well as the novel loss function is generally effective.
\par
The rest of the paper is organized as follows. Section \ref{section:PM} gives the details of the proposed physical-embedded neural network framework, four distinct feature fusion modules and the novel loss function. Section \ref{section: E} presents the experimental results for the models and loss functions on two datasets. We also validate the scalability and computational efficiency of the model. Finally, Section \ref{sectio: CFW} concludes our work and extracts future work.
\begin{figure*}
    \centering
    \includegraphics[width=1\linewidth]{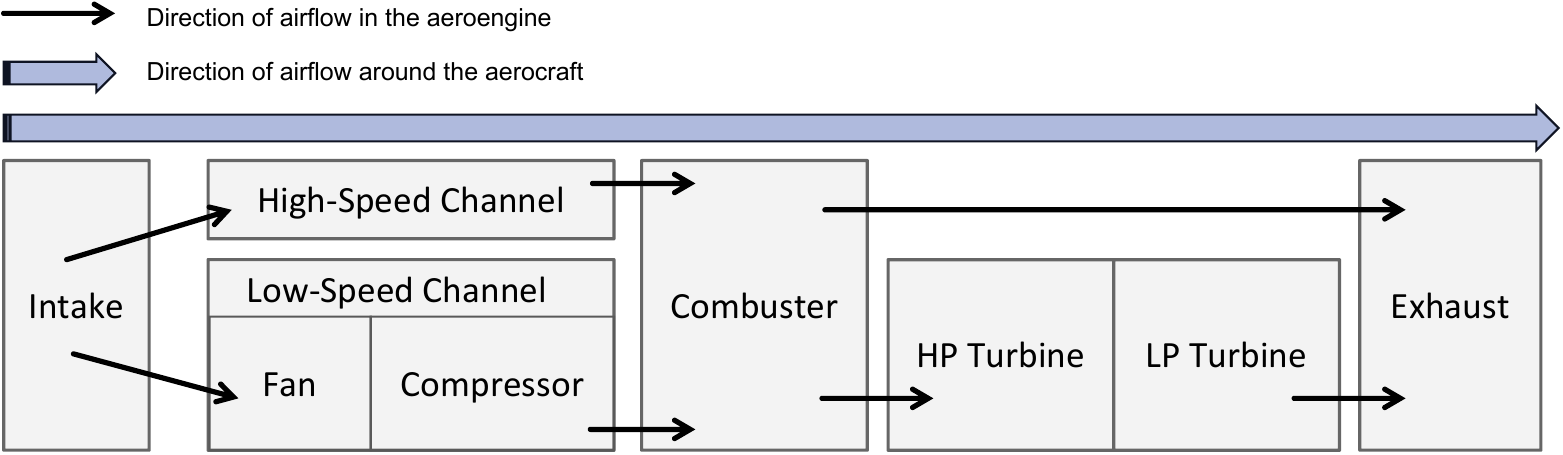}
    \caption{The physical structure of a typical engine}
    \label{fig:coupling}
\end{figure*}
\begin{figure*}[h]
    \centering
    \includegraphics[width=1\linewidth]{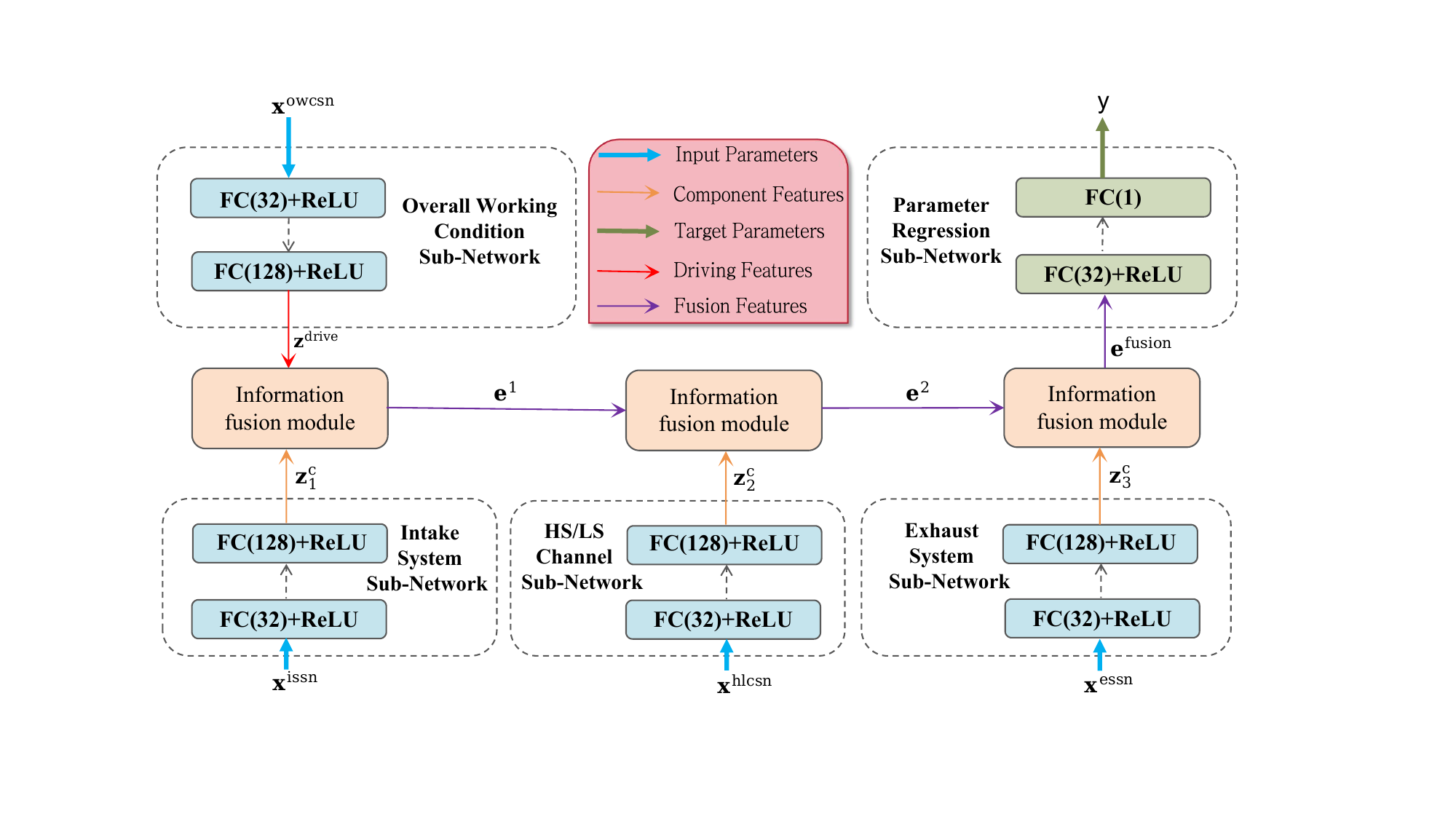}
    \caption{Model architecture of the proposed physical-embedded neural network}
    \label{fig:Architecture}
\end{figure*}

\section{PREDICTION MODELS}
\label{section:PM}
In this section, we will first introduce the proposed model architecture based on the physical structure of a real aeroengine and the technology of neural networks. As the core of the proposed method, four novel information fusion modules are elaborately designed and presented in detail in the following subsection. Finally, we will introduce an innovative loss function for optimizing network parameters more effectively during  the training process.
\subsection{Model Architecture by Fusing Engine Structure and Neural Network}
Data-driven methods have been previously used for aeroengine performance prediction, but most of them are based on generalized neural networks and do not account for the characteristics of the aeroengine structure. This can result in poorly interpretable models with large parameter counts and low accuracy. Instead, we embed the physical structure of the aeroengine as shown in Figure \ref{fig:coupling} into the architectural design of the prediction model. Specifically, we construct sub-networks for the main components separately, and then connect them according to the physical coupling relationships between the aeroengine components. As a result, a novel model architecture for predicting aeroengine performance parameters is formed, which is termed as Physical-Embedded Neural Network (PENN). As shown in Figure \ref{fig:Architecture}, PENN follows the following domain knowledge thoughts of the aeroengine structure and neural networks:
\begin{enumerate}
    \item Each sub-network represents a system composed of real engine components.
    \item The component-level sub-networks are not independent of each other, but rather depend on and interact with one another. The coupling relationships between the sub-networks are consistent with the coupling relationships of the component systems in the real aeroengine.
    \item The data flow and sequence of the entire network are consistent with the real aeroengine. For example, the properties of the airflow are determined by the overall conditions (such as altitude, speed, and air pressure), and the airflow will first enter the aeroengine through the Intake System, then flow through the Fan, the High-Speed/Low-Speed (HS/LS) Channel, and finally out through the Exhaust System. Therefore, the data flow direction of the model should also be consistent with this sequence.
    \item The feature information is fused along the flow. The overall working condition feature is used as the driving feature, which enters the model first. As the feature propagates towards the tail end of the model, it passes through information fusion modules that gradually absorb the supplementary features from other component sub-network, acquiring a group of enhanced features with rich information. 
\end{enumerate}

Mathematically, given the input parameter vector \(\mathbf{x}^\text{in}\), we first divide it into four sets of parameters \(\mathbf{x}^\text{in}=\{\mathbf{x}^\text{owcsn}, \mathbf{x}^\text{isnn}, \mathbf{x}^\text{hlcsn}, \mathbf{x}^\text{essn}\}\), which serve as inputs for the Overall Working Condition Sub-Network (OWCSN), the Intake System Sub-Network (ISSN), the HS/LS Channel Sub-Network (HLCSN), and the Exhaust System Sub-Network (ESSN), respectively. Then, the above four sets of parameters are input into the corresponding component sub-networks to generate the driving feature \(\mathbf{z}^\text{drive}\) and the three component features $\mathbf{z}_\text{1}^\text{c}$,\(\mathbf{z}_\text{2}^\text{c}\),$\mathbf{z}_\text{3}^\text{c}$:
\begin{equation}
\mathbf{z}^\text{drive}=f^\text{owcsn}_\text{mlp}(\mathbf{x}^\text{owcsn})
\end{equation}
\begin{equation}
\mathbf{z}_\text{1}^\text{c}=f^\text{issn}_\text{mlp}(\mathbf{x}^\text{issn})
\end{equation}
\begin{equation}
\mathbf{z}_\text{2}^\text{c}=f^\text{hlcsn}_\text{mlp}(\mathbf{x}^\text{hlcsn})
\end{equation}
\begin{equation}
\mathbf{z}_\text{3}^\text{c}=f^\text{essn}_\text{mlp}(\mathbf{x}^\text{essn})
\end{equation}

\begin{equation}
f_\text{mlp}^\text{*}(\mathbf{x}^\text{*})=ReLU(FC^\text{128}(ReLU(FC^\text{32}(\mathbf{x}^{*}))))
\label{equition: component network}
\end{equation}
where \(FC^\text{32}()\),\(FC^\text{128}()\),\(ReLU()\) are fully connected layer with 32 nodes, fully connected layer with 128 nodes and the popular Rectified Linear Unit (ReLU) activation function, respectively. As can be seen in equition \ref{equition: component network}, the component sub-networks are constructed as multi-layer perceptrons consisting of two consecutive layers, which abstract the corresponding inputs to capture the characteristics of the respective engine component systems. It should be noted that the input dimensionality of the first layer is contingent upon the sensor parameters, which are derived from the simulation data of a real aeroengine. The range of input dimensions for this first layer spans from 3 to 11 for our datasets. Subsequent to the two-layer dimension expansion process, the component sub-networks output the driving feature or the component features all with 128 dimensions. The component sub-networks adhere to the principle of dimensional consistency in the feature space, which makes the following information fusion module to share the network structure possible, thus enhancing the flexibility and scalability of the overall model architecture.
\par
When the driving feature $\mathbf{z}^\text{drive}$ and the three component features $\mathbf{z}_\text{1}^\text{c}$,\(\mathbf{z}_\text{2}^\text{c}\),$\mathbf{z}_\text{3}^\text{c}$ are ready, information fusion modules start with $\mathbf{z}^\text{drive}$ and gradually absorb supplementary information from $\mathbf{z}_\text{1}^\text{c}$,\(\mathbf{z}_\text{2}^\text{c}\),$\mathbf{z}_\text{3}^\text{c}$ to obtain the final fused feature $\mathbf{e}^\text{fusion}$:
\begin{equation}
\mathbf{e}^\text{1}=f_{\text{fusion}}^{1}(\mathbf{z}^\text{drive},\mathbf{z}_{1}^{c})
\end{equation}
\begin{equation}
\mathbf{e}^\text{2}=f_{\text{fusion}}^{2}(\mathbf{e}^\text{1},\mathbf{z}_\text{2}^\text{c})
\end{equation}
\begin{equation}
\mathbf{e}^\text{fusion}=f_{\text{fusion}}^{3}(\mathbf{e}^\text{2},\mathbf{z}_\text{3}^\text{c})
\end{equation}
where $f_{\text{fusion}}^\text{1}$(), $f_{\text{fusion}}^\text{2}$(), $f_{\text{fusion}}^\text{3}$() are information fusion modules. The output feature dimension of the information fusion module is kept consistent with that of the driving feature and the three component features, which ensures the reusability of the information fusion module. Thus, we share the same network structure but learn different network parameters for all information fusion modules. During the fusion process, the information fusion module continuously extracts the valid information output from the component sub-networks and incrementally incorporates it into the overall working condition features. This process is also based on the principles of a real aeroengine to build the coupling relationship between the component systems. Grounded on the theory of multilayer perceptual machines and the principles of the attention mechanism, we have designed four representative information fusion modules, namely Fully-Connected Fusion (FCF), Bottle-Neck Fusion (BNF), Attention-Based Fusion (ABF) and Channel-Adaptive-Weighted Fusion (CAWF), which will be detailed in the next subsection. Correspondingly, prediction models with the above four representative information fusion modules are denoted as PENN-FCF, PENN-BNF, PENN-ABF and PENN-CAWF, respectively.
\par
Finally, the acquired fused feature $\mathbf{e}^\text{fusion}$ is input into the parameter regression sub-network to get the predicted value $\mathit{y}$ for the target parameter such as thrust or specific impulse:
\begin{equation}
    \mathit{y}=FC^\text{1}(ReLU(FC^\text{32}(\mathbf{e}^\text{fusion})))
\end{equation}
where $FC^\text{1}()$ is fully connected layer with 1 output node. As can be seen, the parameter regression sub-network is also realized as multi-layer perceptrons, which gradually compressed the features and eventually downgraded to one-dimensional predicted output value.
\par
From the above description, PENN differs significantly from previous data-driven approaches consuming the input parameters all at once. In contrast, the proposed framework is built based on the actual aeroengine structure and each component sub-network only consumes the sensor parameters of the corresponding component following an information fusion module to built the relationships between adjacent components as well as the overall working condition. As a result, PENN can effectively reduce the number of network parameters and increase the interpretability of the prediction model.

\subsection{Elaborately Designed Information Fusion Modules}
The information fusion module constitutes the core of PENN and represents the pivotal technology that determines the network’s performance. There are two distinct types of inputs to the information fusion module. The driving feature $\mathbf{z}^\text{drive}$ or upper-level fusion feature $\mathbf{e}^\text{1}$/$\mathbf{e}^\text{2}$ serves as the module’s main input, while the component features $\mathbf{z}^\text{c}_\text{1}$/$\mathbf{z}^\text{c}_\text{2}$/$\mathbf{z}^\text{c}_\text{3}$ act as the module’s supplementary input. We employ multiple information interactions or fusion techniques to process the main and supplementary input features, thereby obtaining more comprehensive and enriched fusion features.

\begin{figure*}
    \centering
    \includegraphics[width=1\linewidth]{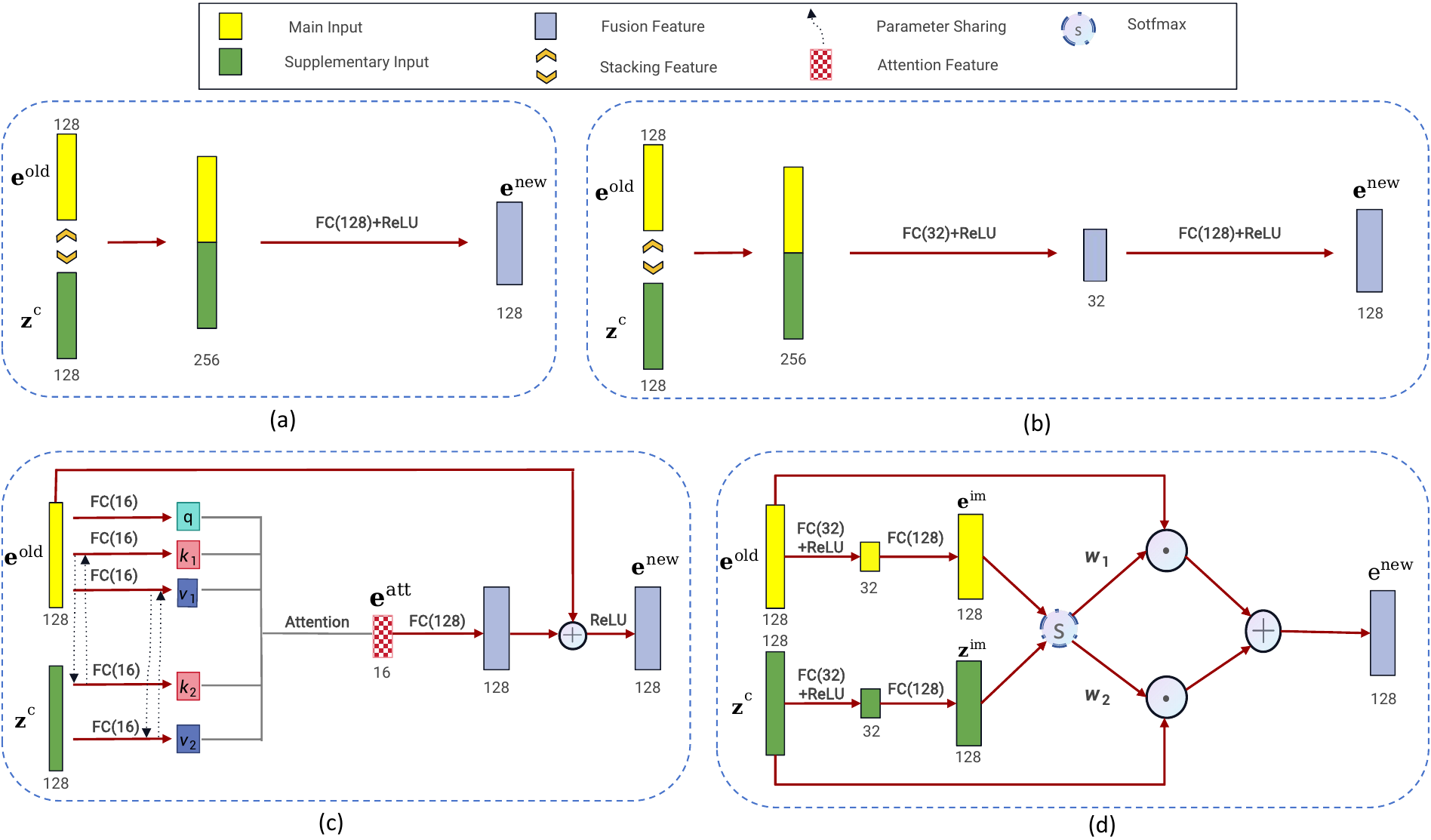}
    \caption{Diagrams of the proposed information fusion modules}
    \label{fig:fusion module}
\end{figure*}

For simplicity, we abbreviate the main input and supplementary input as $\mathbf{e}^\text{old}$ and $\mathbf{z}^\text{c}$, respectively, and the output of the information fusion module as $\mathbf{e}^\text{new}$ in this subsection. The followings are the details of the four elaborately designed information fusion modules as illustrated in Figure \ref{fig:fusion module}.

\subsubsection{Fully-Connected Fusion}

Fully-Connected Fusion (FCF), drawing inspiration from the theory of multi-layer perceptrons, represents the most straightforward approach for feature fusion. It employs a simple fullyconnected layer to achieve feature fusion and dimensionality reduction, yielding information-rich and dimensionally consistent fusion features.
\par
As illustrated in the Figure \ref{fig:fusion module} \textit{(a)}, the main $\mathbf{e}^\text{old}$ and the supplementary input $\mathbf{z}^\text{c}$ are firstly stacked to obtain a 256 dimensional tensor, which is then compressed by a fully-connected layer to acquire the fusion feature $\mathbf{e}^\text{new}$:
\begin{equation}
    \mathbf{e}^\text{new}={ReLU}(FC^\text{128}(
        \mathbf{e}^\text{old} \cup \mathbf{z}^\text{c}
    ))
\end{equation}
where $\cup$ denotes tensor stacking. The strength of FCF lies in its ability to learn intricate information interactions and relationships with a remarkably simple architecture. However, its drawback stems from the considerable increase in the overall number of parameters due to the dense interconnections, leading to computational inefficiency. Furthermore, it is susceptible to the overfitting problem when there are not enough training samples available.

\subsubsection{Bottle-Neck Fusion}
\begin{figure}[h]
    \centering
    \includegraphics[width=0.8\linewidth]{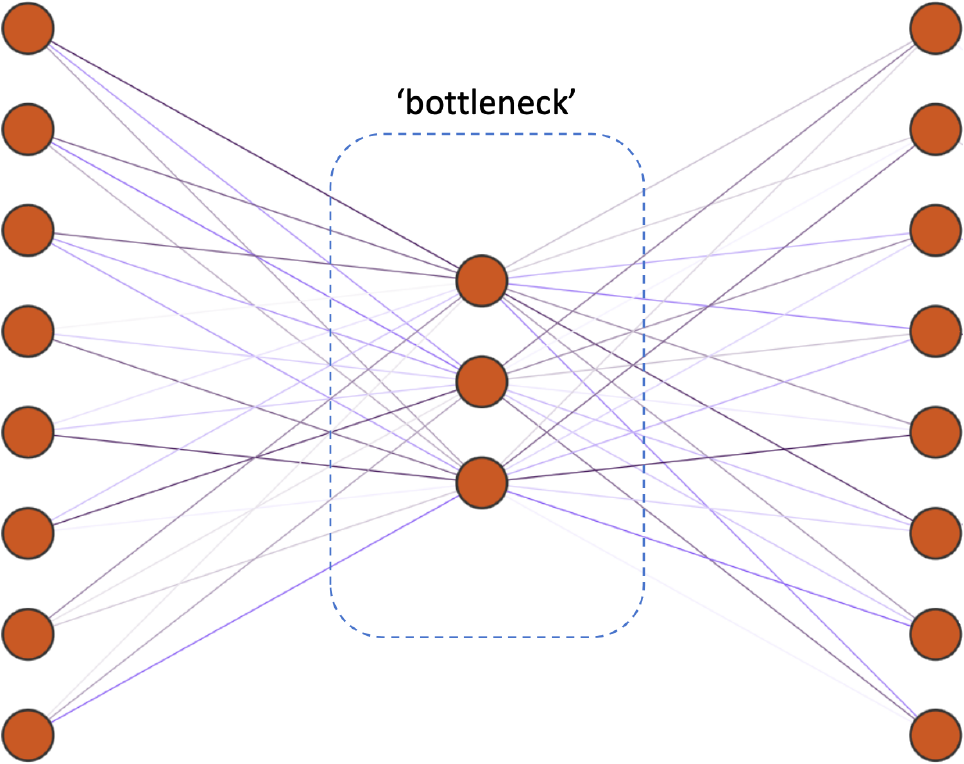}
    \caption{Illustration of bottleneck structure}
    \label{fig:bottle neck}
\end{figure}

A bottleneck structure in a neural network, as illustrated in Figure \ref{fig:bottle neck}, is a layer with fewer neurons compared to the layers preceding and succeeding it. This configuration creates a "bottleneck" that compels the network to learn a compressed representation of the input data. It has been widely applied in fields such as information compression \cite{cho2014properties}, image segmentation \cite{badrinarayanan2017segnet,chen2018encoder,badrinarayanan2015segnet}, and object detection \cite{ji2021cnn,koh2020concept}. Inspired by the success of bottleneck structure, we design the Bottle-Neck Fusion (BNF) module as shown in Figure \ref{fig:fusion module} \textit{(b)}. BNF first converts the stacked tenser of $\mathbf{e}^\text{old}$ and $\mathbf{z}^\text{c}$ to a low dimensional tenser, and then lift its dimension to get the fusion feature $\mathbf{e}^\text{new}$:
\begin{equation}
    \mathbf{e}^\text{new}=ReLU(FC^\text{128}(ReLU(FC^\text{32}(
        \mathbf{e}^\text{old} \cup \mathbf{z}^\text{c}))))
\end{equation}
\par
Compared to FCF, BNF offers three main advantages. Firstly, it effectively reduces the number of network parameters (approximately decreased from $256\times128$ to $256\times32+32\times128$), thereby improving the overall computational efficiency of the model. Secondly, BNF has more network layers than FCF, which increases the depth of the network and enhances the ability to learn more abstract features. Finally, from an information theory perspective, the bottleneck layer may bring potential information loss due to the lower dimensionality. However, this information loss can be beneficial, as it forces the network to learn a more generalized and robust representation of the data, thus can improve the generalization ability of the prediction model across various tasks.

\subsubsection{Attention-Based Fusion}
The attention mechanism, as the core of Transformers \cite{vaswani2017attention}, has been widely applied in natural language understanding \cite{galassi2020attention}, computer vision \cite{han2022survey,arnab2021vivit,10450733,10490963}, multi-modal information fusion \cite{wang2022msran,wang2022amfnet}, and various large foundation models  \cite{derose2020attention}. The key idea is to let the network learn to focus on the most relevant features for the current context or task. This is done by computing attention weights that indicate the importance of each feature, and then taking a weighted sum to fuse the features together.Compared to simple fusion operations like concatenation or addition, attention allows for a more flexible and dynamic combination that adapts based on the content of input data. But at the same time, the attention mechanism also has drawbacks of high computational complexity and requiring a large amount of training data. We investigate it here mainly because of its high popularity.
\par
Specifically, the proposed Attention-Based Fusion (ABF) module, as illustrated in the Figure \ref{fig:fusion module} (c), first maps the main input $\mathbf{e}^\text{old}$ to low-dimensional Query space and the supplementary input $\mathbf{z}^\text{c}$ to low-dimensional Key and Value spaces for learning transformation spaces and reducing computational complexity:
\begin{equation}
\mathbf{q}=FC^\text{16}_\text{query}(\mathbf{e}^\text{old})
\end{equation}
\begin{equation}
\{\mathbf{k}_\text{1},\mathbf{k}_\text{2}\}=FC^\text{16}_\text{key}(\{\mathbf{e}^\text{old},\mathbf{z}^\text{c}\})
\end{equation}
\begin{equation}
\{\mathbf{v}_\text{1},\mathbf{v}_\text{2}\}=FC^\text{16}_\text{value}(\{\mathbf{e}^\text{old},\mathbf{z}^\text{c}\})
\end{equation}
where $FC^\text{16}_\text{query}()$, $FC^\text{16}_\text{key}()$, $FC^\text{16}_\text{value}()$ are three independent fully-connected layers with 16 nodes to implement space transformations; and $\mathbf{q}$, $\mathbf{k}_\text{1}/\mathbf{k}_\text{2}$, $\mathbf{v}_\text{1}/\mathbf{v}_\text{2}$ are the acquired query feature, key features and value features, respectively. Note that only the main input generates query feature for aggregation.
\par
Subsequently, based on the principle of the attention mechanism, the attention tensor $\mathbf{e}^\text{att}$ can be computed by: 
\begin{equation}
\label{equation: Attention_quation}
\mathbf{e}^\text{att}=f^\text{10}_\text{softmax}( \frac{[\mathbf{q}\mathbf{k}_\text{1}^\text{T},\mathbf{q}\mathbf{k}_\text{2}^\text{T}]}{\sqrt{d}})\mathbf{\cdot} \begin{bmatrix} \mathbf{v}_\text{1} \\ \mathbf{v}_\text{2} \end{bmatrix} \\
\end{equation}
where $f^\text{10}_\text{softmax}()$ is the widely used SoftMax function with temperature value of 10, and d is the dimensionality of $\mathbf{q}$.
\par
Finally, we lift the dimension of $\mathbf{e}^\text{att}$ to get the 128- dimensional supplementary feature, which acts as the residual item of the fusion feature $\mathbf{e}^\text{new}$:
\begin{equation}
    \mathbf{e}^\text{new}={ReLU}(\mathbf{e}^\text{old}+FC^\text{128}\mathbf(\mathbf{e}^\text{att}))
\end{equation}
where the function $ReLU()$ is used to restore the non-negativity of the input feature, and the residual structure is plugged in for reducing the information loss of the main input and accelerating the convergence speed of the network.

Unlike in FCF and BNF where the main input and supplementary input have equal status, in ABF, the main input draws useful information from the supplementary input to enrich the main input, which is particularly in line with the design intention of the proposed PENN. However, ABF is susceptible to the insufficient training issue due to insufficient samples.
\subsubsection{Channel-Adaptive-Weighted Fusion}
The Channel-Adaptive-Weighted Fusion (CAWF) module draws on the core ideas of the channel weighting mechanism, which is a variant of the attention mechanism widely applied in the field of computer vision\cite{xu2021adaptive,gao2019change}. As illustrated in the Figure \ref{fig:fusion module} (d), the channel weighting mechanism assigns adaptive weights based on the content of the inputs to every feature channel when combining information from multiple sources. During the fusion process, important features related to the target characteristics are highlighted, while interfering features not related to the target characteristics are suppressed.

Inspired by the famous SENet \cite{hu2018squeeze}, CAWF firstly employs bottleneck structures for mapping the inputs to their corresponding importance tensors $\{\mathbf{e}^\text{im}, \mathbf{z}^\text{im}\}$, which are then normalized to get the weight tensors $\{\mathbf{w}_\text{1}, \mathbf{w}_\text{2}\}$:
\begin{equation}
\label{concentrated feature}
\begin{split}
\{\mathbf{e}^\text{im},\mathbf{z}^\text{im}\}=FC^\text{128}(ReLU(FC^\text{32}(\{\mathbf{e}^\text{old},\mathbf{z}^\text{c}\})))
\end{split}
\end{equation}

\begin{equation}
\begin{split}
    \{\mathbf{w}_\text{1}, \mathbf{w}_\text{2}\} = f^\text{1}_\text{softmax\_c}(\{\mathbf{e}^\text{im}, \mathbf{z}^\text{im}\}) 
\end{split}
\label{eq:softmax}
\end{equation}
where $f_\text{softmax\_c}^\text{1}()$ is the SoftMax function along the channel dimension with temperature value of 1. The acquired weight tensors reflect the relative importance of each feature in the corresponding dimension.
\par
 Finally, CAWF utilizes these obtained weight tensors to perform a weighted summation of the two input features across each feature dimension, resulting in the 128-dimensional final fusion features $\mathbf{e}^\text{new}$:

\begin{equation}
\mathbf{e}^\text{new} = \mathbf{w}_\text{1} \odot \mathbf{e}^\text{old} + \mathbf{w}_\text{2} \odot \mathbf{z}^\text{c}
\end{equation}
where $\odot$ denotes element-wise multiplication.
Similar to ABF, CAWF can learn the importance of information base on its content during the learning process. However, CAWF is realized based on fully-connected structures, which avoids the complex computation of the attention mechanism and significantly improves the computational efficiency. Compared with the traditional feature weighting idea, CAWF realizes the independent adaptive weighting of each feature dimension, resulting in higher degree of fusion between the features. In addition, CAWF get the weight tensors by comparing the importance of the inputs, which reflects the idea that the strength and weakness are relative.

\subsection{Loss Function}
The training of the model is a crucial aspect in the development of the PEEN model. Model training for neural networks is fundamentally an optimization problem, where the goal is to minimize a loss function by continuously adjusting the model parameters. This gradual reduction in the loss function leads to an improvement in the model’s predictive capabilities.
\par
For regression tasks, popular loss functions are Mean Squared Error (MSE) and Mean Absolute Error (MAE):
\begin{equation}
L_\text{MSE} = \frac{1}{n}\sum_{i=1}^n (y_i - \hat{y}_i)^2
\end{equation}
\begin{equation}
L_\text{MAE} = \frac{1}{n}\sum_{i=1}^n |y_i - \hat{y}_i|
\end{equation}
where $n$ is the number of samples, $y_i$ is the ground-truth value, and $\hat{y}_i$ is the predicted value.
\par
Obviously, $L_\text{MSE}$ and $L_\text{MAE}$ are sensitive to the absolute values of predicted parameters which can lead to relatively large prediction errors for samples with small ground-truth values. However, we usually pay more attention to relative errors in practical applications. To address this issue, a new loss function, termed as Mean Absolute Relative Error (MARE) $L_\text{MARE}$, is proposed as follows:
\begin{equation}
\label{MARE}
L_\text{MARE} = \frac{1}{n}\sum_{i=1}^n \left|\frac{y_i - \hat{y}_i}{y_i}\right|
\end{equation}
\par
As can be seen, $L_\text{MARE}$ calculates the absolute relative differences between the predicted and ground-truth values, rather than the absolute differences. Thus, $L_\text{MARE}$ can normalize the errors for different absolute values, and can achieve more balanced optimization effects for both large-valued samples and small-valued samples.

\section{EXPERINMENT}

In this section, we will first introduce the datasets, the baseline models and the experimental settings. Next, the proposed hybrid model and the novel loss function are validated by extensive datasets experiments. Finally, we validate the scalability of the model and displays its computational efficiency to provide reference for practical applications.
\label{section: E}
\subsection{Datasets}

The simulation data of two typical aeroengines, namely High-Speed Dataset and Low-Speed Dataset, are employed for training and testing the proposed hybrid model as well as the novel loss function. Both datasets have 18 input parameters and 2 output parameters which will be detailed in Subsection \ref{experimental settings}.

\textbf{High-Speed Dataset.} A series of data generated during the random envelope flight simulation process of a certain combination engine above Mach 2, with a total of 50,000 experimental samples. We randomly selected 30,000 and 10,000 samples to form the training set and validation set respectively, and the remaining 10,000 samples formed the test set.
\par
\textbf{Low-Speed Dataset.} A series of simulation data was  generated during the random envelope flight below Mach 2.3, with a total of 20,000 experimental samples. We randomly selected 12,000 and 4,000 samples to form the training set and validation set respectively, and the remaining 4,000 samples formed the test set. It should be noted that this dataset contains a few samples with zero-value impulse making it impossible to train the model when using $L_\text{MARE}$. After analysis, it was found that negative thrust and zero-value impulse always occur at the same time. Therefore, we eliminated samples with zero-value impulse in the dataset and directly set it as zero when the predicted thrust is negative.

\subsection{Baseline}
\begin{figure}[h]
    \centering
    \includegraphics[width=1\linewidth]{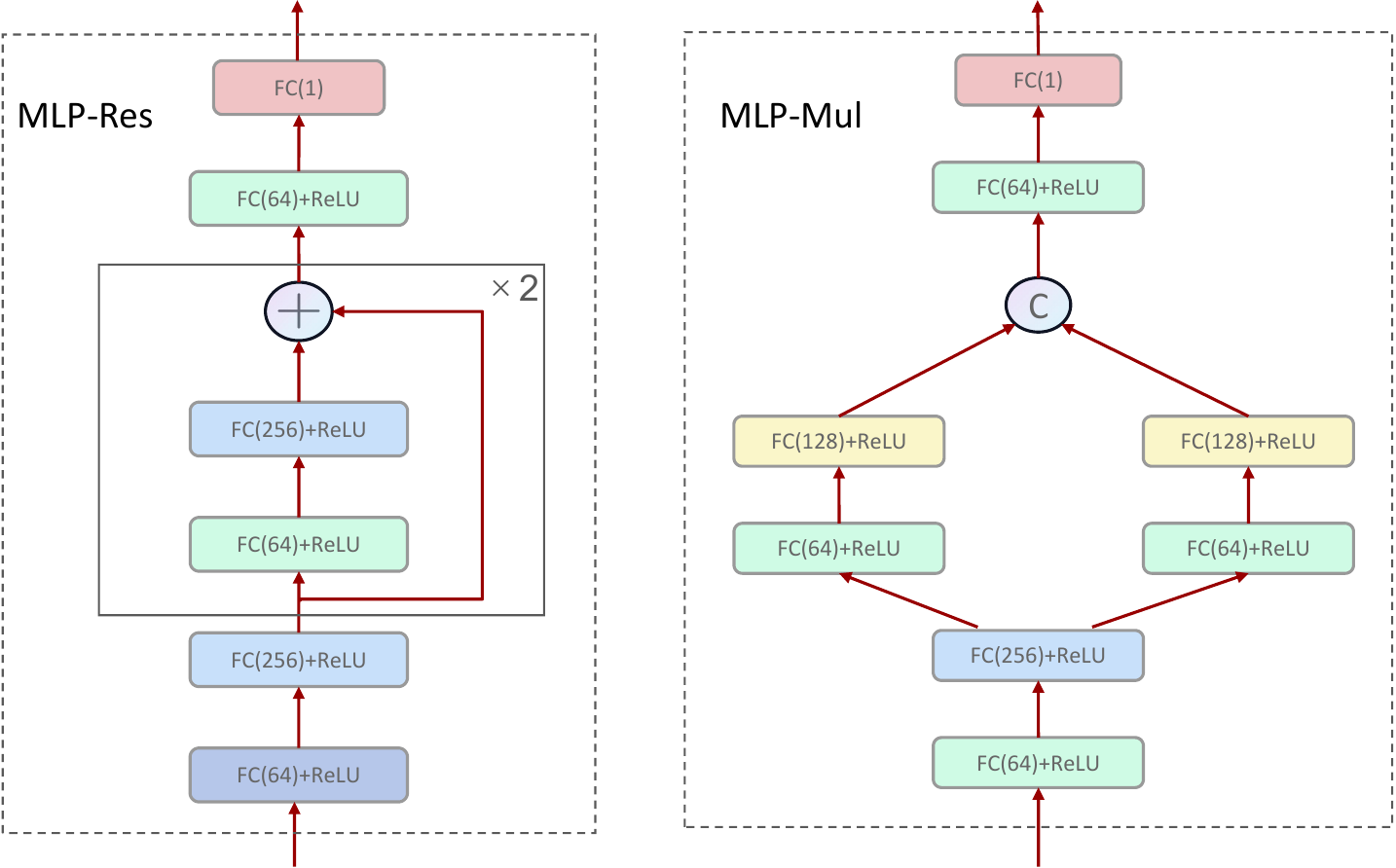}
    \caption{Baseline models}
    \label{fig:twoMLP}
\end{figure}

We have totally designed four MLP models with different network structures and depths. For brevity, we only choose two most competitive models of them as the baseline model, namely MLP-Res and MLP-Mul as shown in Figure \ref{fig:twoMLP}. 

\textbf{MLP-Res} is an 8-layer architecture built upon the standard fully-connected structure, with the addition of two residual modules. The residual module consists of a skip connection and a bottleneck structure. Compared to a regular MLP, this approach helps to alleviate the vanishing gradient problem, thereby improving the network’s convergence speed and learning capability.

\textbf{MLP-Mul} is a 6-layer MLP network incorporating an extra parallel processing branch. In the computer vision domain, leveraging parallel multi-branch structures has been demonstrated to be effective for fusing information from different perspectives \cite{weng2022deep,he2020multi}. In this study, the multi-branch structure can extract features from different subspaces, enabling the decoupling of feature groups. This enhancement enhances the network’s stability and overall performance.

\begin{table*}[h]
\centering
\caption{Input and Output Parameters for PENN}
\begin{tabular}{l|p{7cm}|p{3cm}}
    \textbf{Parameter Group} & \textbf{Input Parameters} & \textbf{Output Parameters} \\
    \hline
    \multicolumn{1}{c|}{\textbf{Overall Parameters}} & 
    Atmospheric Static Pressure, \newline
    Atmospheric Static Temperature, \newline
    Aircraft Flight Mach Number &
    Thrust, \newline
    Specific Impulse \\
    \cline{1-2}
    \multicolumn{1}{c|}{\textbf{Intake System}} & 
    Total Pressure Recovery Coefficient of Intake, \newline
    Total Mass Flow Rate of Intake &
     \\
    \cline{1-2}
    \multicolumn{1}{c|}{\textbf{Low Speed Channel}} &
    Relative Physical Speed of Fan, \newline
    Relative Physical Speed of Compressor, \newline
    Total Temperature at Low Pressure Turbine Outlet, \newline
    Total Pressure at Fan Inlet, \newline
    Static Pressure at Low Pressure Turbine Outlet, \newline
    Engine Pressure Ratio, \newline
    Turbine Throttle Lever Angle, \newline
    Relative Opening of Turbine Outlet &
     \\
    \cline{1-2}
    \multicolumn{1}{c|}{\textbf{High Speed Channel}} & 
    Area Ratio of Bypass Duct Outlet to Mixing Chamber, \newline
    Combined Engine Throttle Angle, \newline
    Fuel Flow Rate of Ramjet Combustion Chamber &
     \\
    \cline{1-2}
    \multicolumn{1}{c|}{\textbf{Exhaust System}} &
    Throat Area of Nozzle, \newline
    Outlet Area of Nozzle Expansion Section &
     \\
\end{tabular}
\label{tab:parameters_PENN}
\end{table*}
\subsection{Experimental Settings}
\label{experimental settings}
We now give the detail of experimental settings such as the training design, the input and output design, and the performance metrics.

\textbf{Trainning Design.} For the sake of fairness, we kept most of the experimental settings consistent across different prediction models. We used the Adam optimizer \cite{Kingma2014AdamAM} with a total of 150 training epochs. The batch size is set as 100 and 40 for the High-Speed dataset(HS) and the Low-Speed dataset(LS), respectively. The initial learning rate is set as 0.01 for MLP models, and is reduced by a factor of 10 at the 80th and 120th epochs. As for PENN models, the initial learning rate is set as 0.002, and is reduced by a factor of 2 at the 60th, 80th, and 100th epochs. The discrepancy in milestone configurations for learning rate scheduling can be attributed to the inherent complexity of PENN architectures in comparison to MLP networks, necessitating an earlier reduction in the learning rate to facilitate effective training convergence. All the experiments are carried out on a PC with AMD Ryzen™ 9 CPUs, 64-GB RAM, and GEFORCE GTX 3090 GPUs.

\textbf{Input and Output Design.}
 A total of 18 input parameters are recorded during the simulation of a real aeroengine, which are further divided into five groups as shown in TABLE \ref{tab:parameters_PENN}. The output parameters are thrust and specific impulse. For PENN, we merge high-speed and low-speed parameters as the input for HLCSN, and other component sub-networks consume their corresponding input parameters. For MLPs, we directly feed all 18 input parameters into the model.

\textbf{Performance Metric.}
 For regression models, the Mean Absolute Percentage Error(MAPE) is generally used as the evaluation index, which is defined as follows:
\begin{equation}
\text{MAPE} = \frac{1}{n} \sum_{i=1}^{n} \left| \frac{A_i - F_i}{A_i} \right| \times 100\%
\end{equation}
where $n$ denotes the number of samples in the test set, $A_i$ is the ground-truth value, and $F_i$ is the predicted value. The smaller the value of MAPE, the better the model performance, and vice versa. Besides MAPE, we also compare model parameter quantity, which can usually reflect the computational efficiency and can be approximately calculated as: 
\begin{equation}
\text{P}=\sum_{l=1}^{L-1} (n_l + 1) \cdot n_{l+1}
\end{equation}
where $L$ denotes the total number of layers in the neural network, including the input and output layers. $n_l$ represents the number of nodes (neurons) in the $l$-th layer.
\begin{table*}[h]
\centering
\caption{Comparation of prediction errors for different models on two datasets}
\begin{threeparttable}
\setlength{\tabcolsep}{3mm}{
\begin{tabular}{*{11}{c}}

  \toprule
  \multirow{2}*{\textbf{Prediction}} & \multirow{2}*{\textbf{Params}} & \multicolumn{3}{c}{\textbf{HS dataset}} & \multicolumn{3}{c}{\textbf{LS dataset}} & \multicolumn{3}{c}{\textbf{Synthesis}\tnote{1}}\\  
  \cmidrule(lr){3-5}\cmidrule(lr){6-8}\cmidrule(lr){9-11}\morecmidrules\cmidrule(lr){9-11}
  \textbf{model}&& Thrust & Impulse & Average & Thrust & Impulse & Average & Thrust & Impulse & Average\\
  \midrule
  MLP-Res& 100k&0.31\%&0.31\% & \textbf{0.31\%} & 2.78\% & 3.25\% & 3.01\% & 1.55\% & 1.78\% & 1.66\%\\
  MLP-Mul& 84k & 0.35\% & 0.34\% & 0.35\% & 2.59\% & 2.61\% & 2.60\% & 1.47\% & 1.48\% & 1.48\%\\
  \cmidrule(lr){0-1}
  PENN-FCF& 120k & 0.31\% & 0.54\% & 0.43\% & 2.10\% & 2.52\% & 2.31\% & 1.21\% & 1.53\% & 1.37\%\\
  PENN-BNF& 59k & 0.25\% & 0.40\% & 0.33\% & 2.37\% & 2.31\% & 2.34\% & 1.31\% & 1.36\% & 1.34\%\\  
  PENN-ABF\tnote{2}&46k & 0.64\% & 0.65\% & 0.65\% & 3.02\% & 3.73\% & 3.38\% & 1.83\% & 2.19\% & 2.02\%\\
  PENN-CAWF&71k & 0.25\% & 0.54\% & 0.40\% & 2.21\% & 2.25\% & \textbf{2.23\%} & 1.23\% & 1.40\% & \textbf{1.32}\textbf{\%}\\
  \bottomrule

\end{tabular}}
\begin{tablenotes}    
        \footnotesize               
        \item[1] Taking average over the two datasets (for example: $\text{Thrust}_\text{synthesis}=(  \text{Thrust}_\text{HS} + \text{Thrust}_\text{LS} )/2$ )
        \item[2] PENN-ABF exists a complex attention mechanism computation whose computational efficiency cannot be measured in terms parameter amount       
      \end{tablenotes}            
    \end{threeparttable}
\label{tab:main_result}
\end{table*}

\subsection{Comparative Analysis}
We test the proposed PENN models (PENN-FCF, PENN-BNF, PENN-ABF, PENN-CAWF) and MLP models (MLP-Res, MLP-Mul) on HS and LS datasets, and the prediction results are presented in Table \ref{tab:main_result}. It can be seen that: (1) The errors on LS dataset are obviously larger than those on HS dataset for all prediction models. This is mainly because that the physical laws under LS dataset are more complex than those under HS dataset. (2) From the comprehensive results on HS and LS datasets, PENN-CAWF and PENN-BNF achieved similar optimal prediction performance. This indicates that the physical structure employed by PENN models are effective in extracting and aggregating effective features. (3) Considering the computational efficiencies of the models (except for PENN-ABF), PENN-BNF has the highest computational efficiency, followed by PENN-CAWF. This verifies that the physical structure employed by PENN can effectively reduce invalid network connections, thus enhancing the computational efficiency of the prediction model. (4) PENN-BNF has the best efficiency-performance trade-off among all models, showing the effectiveness of utilizing bottleneck structure in the field of engine performance prediction.
\begin{figure}[h]
    \centering
    \resizebox{0.5\textwidth}{!}{
    \begin{tikzpicture}
        \begin{axis}[
            xlabel={Parameters (10k)},
            ylabel={Mean Comprehensive Error (MAPE \%)},
            grid=major,
            xmin=0,xmax=15,
            ymin=0,ymax=3,
            mark size=3pt,
            legend pos=outer north east
        ]
        \addplot[only marks,mark=square*,color=blue] coordinates {
            (10, 1.66)
        };
        \addplot[only marks,mark=triangle*,color=red] coordinates {
            (8.3, 1.48)
        };
        \addplot[only marks,mark=diamond*,color=green] coordinates {
            (12, 1.37)
        };
        \addplot[only marks,mark=star,color=brown] coordinates {
            (5.8, 1.34)
        };
        \addplot[only marks,mark=pentagon*,color=purple] coordinates {
            (4.6, 2.02)
        };
        \addplot[only marks,mark=oplus,color=orange] coordinates {
            (7.1, 1.32)
        };
        \legend{MLP-Res,MLP-Mul,PENN-FCF,PENN-BNF,PENN-ABF,PENN-CAWF}
        \end{axis}
    \end{tikzpicture}
    }
    \caption{Mean comprehensive error of different models}
    \label{fig:Mean_comprehensive_error}
    \vspace{-0.5cm}
\end{figure}
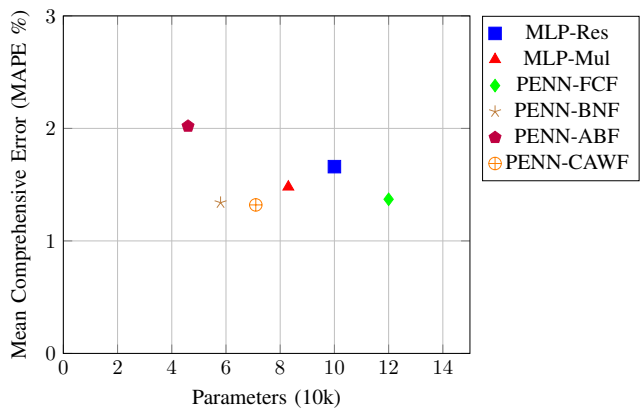
\par

Figure \ref{fig:Mean_comprehensive_error} shows the "error-efficiency" scatter plot of the comprehensive results on HS and LS datasets for all prediction models. The model closer to the bottom left corner in the scatter plot has the best overall performance. From the graph, it can be seen that PENN-BNF has the best trade-off between prediction performance and computational efficiency. In the following more in-depth experiments, we select PENN-BNF as the default prediction model.

\subsection{Validation of Proposed Loss Function}
\begin{table*}[h]
\centering
\caption{Prediction errors under different loss function on two datasets}
\setlength{\tabcolsep}{3mm}{
\begin{tabular}{*{11}{c}}

  \toprule
  \multirow{2}*{\textbf{Loss Function}} & \multirow{2}*{\textbf{Model}} & \multicolumn{3}{c}{\textbf{HS dataset}} & \multicolumn{3}{c}{\textbf{LS dataset}} & \multicolumn{3}{c}{\textbf{Synthesis}\tnote{1}}\\  
  \cmidrule(lr){3-5}\cmidrule(lr){6-8}\cmidrule(lr){9-11}\morecmidrules\cmidrule(lr){9-11}
  & & Thrust & Impulse & Average & Thrust & Impulse & Average & Thrust & Impulse & Average\\
  \midrule
  $L_\text{MSE}$& &0.34\%&0.30\% & \textbf{0.32}\% & 5.35\% & 3.67\% & 4.51\% & 2.85\% & 1.99\% & 2.42\%\\
  $L_\text{MAE}$& MLP-Mul & 0.40\% & 0.40\% & 0.40\% & 3.25\% & 3.19\% & 3.22\% & 1.83\% & 1.80\% & 1.82\%\\
  $L_\text{MARE}$&  & 0.35\% & 0.34\% & 0.35\% & 2.59\% & 2.61\% & \textbf{2.60\%} & 1.47\% & 1.48\% & \textbf{1.48}\textbf{\%}\\
  \midrule
  $L_\text{MSE}$& &0.41\% & 0.46\% & 0.36\% & 5.61\% & 2.69\% & 4.15\% & 3.01\% & 1.58\% & 2.26\%\\
  $L_\text{MAE}$& PENN-BNF & 0.36\% & 0.43\% & 0.40\% & 3.63\% & 2.23\% & 2.93\% & 2.00\% & 1.33\% & 1.67\%\\
  $L_\text{MARE}$& & 0.25\% & 0.40\% & \textbf{0.33}\% & 2.37\% & 2.31\% & \textbf{2.34\%} & 1.31\% & 1.36\% & \textbf{1.34\%} \\
  \bottomrule

\end{tabular}}
\label{tab:loss_function_result}
\end{table*}
\par
To verify the effectiveness of the proposed loss function for different network structures, we choose two representative prediction models, MLP-Mul and PENN-BNF, among the generalized neural network model and the physical-embedded nerual network model as the experimental models. The prediction results for the proposed loss function ($L_\text{MARE}$) as well as two other alternatives ($L_\text{MSE}$, $L_\text{MAE}$) are presented in TABLE \ref{tab:loss_function_result}. It can be seen that when using $L_\text{MSE}$, $L_\text{MAE}$, and $L_\text{MARE}$ respectively, the comprehensive prediction error for PENN-BNF first decreased from 2.26\% to 1.67\%, and finally decreased to 1.34\%, which demonstrates the significant impact of the loss function on predictive performance. In addition, the comprehensive prediction error for MLP-Mul displays the similar trend, which proves that the proposed loss function is universal and can improve the prediction performance under various network architectures.

\subsection{Dataset Size Dependence Experiments}

\begin{figure*}[ht]
\centering
\begin{subfigure}[b]{0.45\textwidth}
\includegraphics[width=\textwidth]{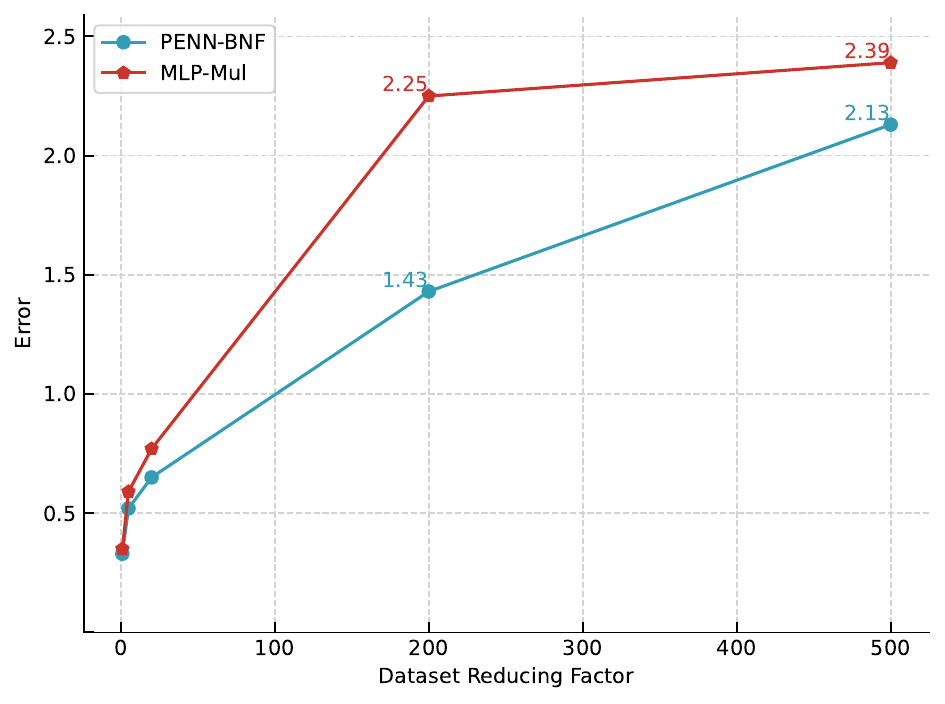}
\label{fig:reduced_HS}
\end{subfigure}
\hfill
\begin{subfigure}[b]{0.45\textwidth}
\includegraphics[width=\textwidth]{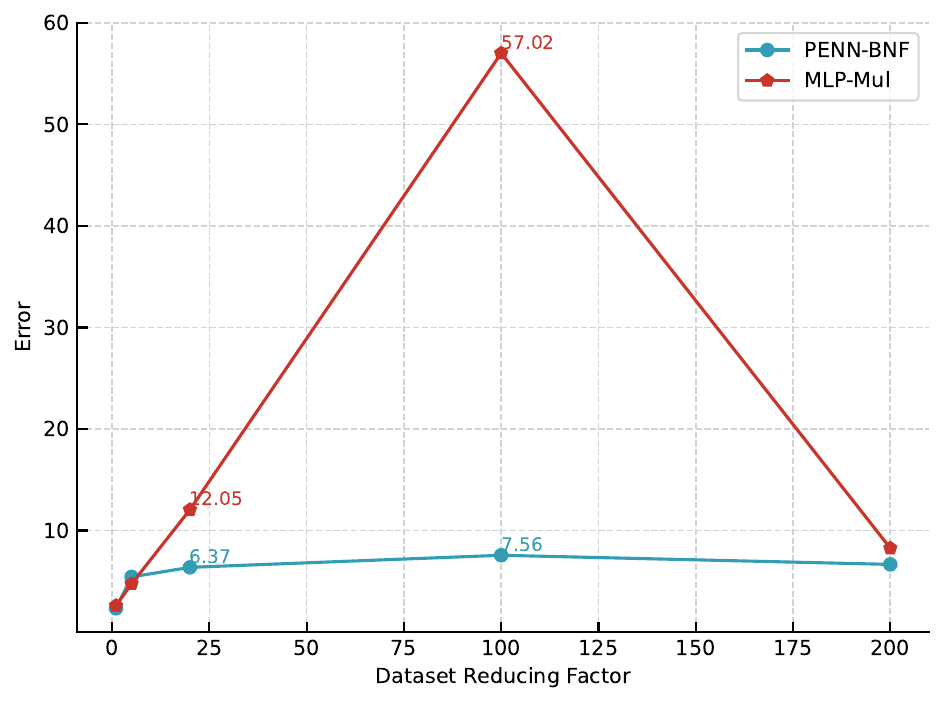}
\label{fig:reduced_LS}
\end{subfigure}
\caption{Prediction errors on scaled HS dataset (left) and scaled LS dataset (right)}
\label{fig:reduced_HS_LS}
\end{figure*}

In order to verify the effectiveness of the proposed physical-embedded neural network structure on different dataset sizes, we scale down the size of the HS and LS datasets by 5, 20, 200, 500 times and 5, 20, 100, 200 times respectively, and use the scaled datasets to train the models. We still choose MLP-Mul and PENN-BNF as the experimental models, and test their prediction performance under different dataset sizes. For the convenience and fairness of the experiment, we did not adjust the hyper-parameters of network training for different dataset sizes. The average prediction errors of thrust and specific impulse under different scaled datasets are recorded and shown as curves in Figure \ref{fig:reduced_HS_LS}. We can see that the average prediction error of PENN-BNF increases gently for both scaled datasets, thus it has stable performance expectations on datasets of different scales. In addition, the average prediction errors of MLP-Mul is always higher than those of PENN-BNF, and MLP-Mul does not even converge when the sampling ratio is 100 on LS dataset. 
We believe that by adjusting hyperparameters as targeted as possible, MLP-Mul in this case can also converge, but PENN-BNF does not require any adjustment. This proves that prediction models with physical-embedded neural network structure is easier to train than those with general neural network structure under different dataset sizes.

\par

\subsection{Model Scalability Experiments}

Due to the constraints of computing resources, we usually need prediction models with different computational complexities in practical applications. Therefore, we obtained PENN-BNF-UP2 and PENN-BNF-UP4 by synchronously increasing the number of nodes in each network layer of PENN-BNF by 2 and 4 times respectively, and obtained PENN-BNF-DOWN2 and PENN-BNF-DOWN4 by synchronously reducing the number of nodes in each network layer of PENN-BNF by 2 and 4 times respectively. The four scaled models mentioned above, together with PENN-BNF, constitute the PENN-BNF family.
\par
We trained and tested each of the five models in PENN-BNF family using HS and LS datasets, and the main results are shown in Table \ref{tab:scalability}. We can see that only when the model parameters is reduced from 59k (PENN-BNF) to 4k (PENN-BNF-DOWN4) does the performance deteriorate significantly. At the same time, scaling up the model can continuously improve predictive performance, but does not yield significant gains. For example, PEEN-BNF-UP4 is only 0.25\% more accurate than PENN-BNF, but its model parameters is 16.7 times that of PENN-BNF and the performance gain brought by the increase in model capacity is relatively small.
\begin{table*}[h]
\centering
\caption{Prediction errors of model in PENN-BNF family}
\begin{threeparttable}
\setlength{\tabcolsep}{2.5mm}{
\begin{tabular}{*{11}{c}}

  \toprule
  \multirow{2}*{\textbf{Model}} & \multirow{2}*{\textbf{Parameters}} & \multicolumn{3}{c}{\textbf{HS dataset}} & \multicolumn{3}{c}{\textbf{LS dataset}} & \multicolumn{3}{c}{\textbf{Synthesis}}\\  
  \cmidrule(lr){3-5}\cmidrule(lr){6-8}\cmidrule(lr){9-11}\morecmidrules\cmidrule(lr){9-11}
  & & Thrust & Impulse & Average & Thrust & Impulse & Average & Thrust & Impulse & Average\\
  \midrule
  PENN-BNF-Down4& 4k&0.67\%&0.65\% & 0.66\% & 6.46\% & 4.53\% & 5.50\% & 3.57\% & 2.60\% & 3.09\%\\
  PENN-BNF-Down2& 15k & 0.41\% & 0.41\% & 0.41\% & 2.90\% & 2.69\% & 2.80\% & 1.66\% & 1.55\% & 1.61\%\\
  \cmidrule(lr){0-1}
  PENN-BNF& 59k & 0.25\% & 0.40\% & 0.33\% & 2.37\% & 2.31\% & 2.34\% & 1.31\% & 1.36\% & 1.34\%\\
  \cmidrule(lr){0-1}
  PENN-BNF-Up2& 233k & 0.23\% & 0.35\% & 0.29\% & 2.07\% & 2.11\% & 2.09\% & 1.15\% & 1.23\% & 1.19\%\\  
  PEEN-BNF-Up4&925k & 0.25\% & 0.28\% & \textbf{0.27\%} & 2.19\% & 1.86\% &\textbf{ 2.03\%} & 1.06\% & 1.11\% & \textbf{1.09\%} \\
  \bottomrule

\end{tabular}}           
    \end{threeparttable}
\label{tab:scalability}
\end{table*}

\par
Overall, PENN-BNF strike a good balance between computational complexity and prediction accuracy, is the advocated solution. For situations where high prediction accuracy is required and computing resources are sufficient, PENN-BNF-UP4 or PENN-BNF-UP2 is a good choice. In situations where computing resources are scarce, PENN-BNF-DOWN2 is also acceptable.
\subsection{Model Efficiency Testing}
We now evaluate the computational efficiency of the five models in PENN-BNF family. We train and run the model on two datasets using Central Processing Units (CUP, AMD Ryzen™ 9 5900X), and the time costs are recorded and shown in Table \ref{tab:efficiency}. It can be seen that the training time and inference time under CPU increase with the size of the models obviously, but the growth times are not the same. The reason may be that the CPU used in the experiment is multi-core and has a certain degree of parallel processing ability. Overall, CPU is sufficient for the models designed in this article. 
\begin{table*}[h]
\centering
\caption{Time costs for different models}
\begin{threeparttable}
\setlength{\tabcolsep}{2mm}{
\begin{tabular}{*{6}{c}}

  \toprule
  \multirow{2}*{\textbf{Model}} & \multicolumn{2}{c}{\textbf{HS dataset}} & \multicolumn{2}{c}{\textbf{LS dataset}} &\\  
  \cmidrule(lr){2-3}\cmidrule(lr){4-5}
  &Training & Inference & Training & Inference \\
  \midrule
  PENN-BNF-Down4&157.90s&0.000154s & 121.64s & 0.000150s \\
  PENN-BNF-Down2 & 172.68s & 0.000167s& 122.91s & 0.000159s \\
  \cmidrule(lr){0-0}
  PENN-BNF & 201.22s & 0.000170s & 143.05s & 0.000164s  \\
  \cmidrule(lr){0-0}
  PENN-BNF-Up2& 260.58s & 0.000189s & 179.95s & 0.000197s  \\  
  PEEN-BNF-Up4& 406.17s & 0.000242s & 281.74s & 0.000259s \\
  \bottomrule

\end{tabular}}         
    \end{threeparttable}
\label{tab:efficiency}
\end{table*}

\section{CONCLUSION AND FUTURE WORK}
\label{sectio: CFW}
Prediction of engine performance is a crucial part for engine design, maintenance, and optimization endeavours. In this study, we design a physical-embedded neural network architecture for real-time prediction of engine performance parameters. In the proposed architecture, the component inputs and internal relationships strictly follow the real engine principles, which greatly increases the model interpretability. As the core of the architecture, four distinct feature fusion modules are deliberately designed and the corresponding prediction models are obtained. Additionally, a novel loss function called mean absolute relative error is proposed to strength the model training effect. Experimental results validate that the proposed architecture can extract and aggregate effective features and reduce invalid network connections, thus simultaneously improving model performance and efficiency. Furthermore, the novel loss function is universal and can improve the prediction performance under various network architectures. We also conduct extensive experiments on datasets scalability, model scalability and efficiency testing, providing a reference for researchers and engineers in different application scenarios.

Moving forward, in order to further improve the model performance and computational efficiency, as well as to increase the model interpretability, we are ready to refine the input information flow and component sub-networks even more. Additionally, we plan to increase the supervision of the sub-networks, which may lead to further enhancements in the model’s capabilities.

\bibliography{IEEE_TAES_regular_template_latex/references}

\begin{thebibliography}{10}
\providecommand{\url}[1]{#1}
\csname url@samestyle\endcsname
\providecommand{\newblock}{\relax}
\providecommand{\bibinfo}[2]{#2}
\providecommand{\BIBentrySTDinterwordspacing}{\spaceskip=0pt\relax}
\providecommand{\BIBentryALTinterwordstretchfactor}{4}
\providecommand{\BIBentryALTinterwordspacing}{\spaceskip=\fontdimen2\font plus
\BIBentryALTinterwordstretchfactor\fontdimen3\font minus \fontdimen4\font\relax}
\providecommand{\BIBforeignlanguage}[2]{{%
\expandafter\ifx\csname l@#1\endcsname\relax
\typeout{** WARNING: IEEEtran.bst: No hyphenation pattern has been}%
\typeout{** loaded for the language `#1'. Using the pattern for}%
\typeout{** the default language instead.}%
\else
\language=\csname l@#1\endcsname
\fi
#2}}
\providecommand{\BIBdecl}{\relax}
\BIBdecl

\bibitem{10.1115/1.4049410}
\BIBentryALTinterwordspacing
A.~M. Briones, A.~W. Caswell, and B.~A. Rankin, ``{Fully Coupled Turbojet Engine Computational Fluid Dynamics Simulations and Cycle Analyses Along the Equilibrium Running Line},'' \emph{Journal of Engineering for Gas Turbines and Power}, vol. 143, no.~6, p. 061019, 03 2021. [Online]. Available: \url{https://doi.org/10.1115/1.4049410}
\BIBentrySTDinterwordspacing

\bibitem{en13112846}
\BIBentryALTinterwordspacing
M.~Ciampolini, S.~Bigalli, F.~Balduzzi, A.~Bianchini, L.~Romani, and G.~Ferrara, ``Cfd analysis of the fuel–air mixture formation process in passive prechambers for use in a high-pressure direct injection (hpdi) two-stroke engine,'' \emph{Energies}, vol.~13, no.~11, 2020. [Online]. Available: \url{https://www.mdpi.com/1996-1073/13/11/2846}
\BIBentrySTDinterwordspacing

\bibitem{osti_1765177}
\BIBentryALTinterwordspacing
``Computational fluid dynamics combustion modeling for rotating detonation engines,'' 1 2021. [Online]. Available: \url{https://www.osti.gov/biblio/1765177}
\BIBentrySTDinterwordspacing

\bibitem{laskowski2016future}
G.~M. Laskowski, J.~Kopriva, V.~Michelassi, S.~Shankaran, U.~Paliath, R.~Bhaskaran, Q.~Wang, C.~Talnikar, Z.~J. Wang, and F.~Jia, ``Future directions of high fidelity cfd for aerothermal turbomachinery analysis and design,'' in \emph{46th AIAA fluid dynamics conference}, 2016, p. 3322.

\bibitem{aerospace10090782}
\BIBentryALTinterwordspacing
Y.~Xu, L.~Gao, R.~Cao, C.~Yan, and Y.~Piao, ``Power balance strategies in steady-state simulation of the micro gas turbine engine by component-coupled 3d cfd method,'' \emph{Aerospace}, vol.~10, no.~9, 2023. [Online]. Available: \url{https://www.mdpi.com/2226-4310/10/9/782}
\BIBentrySTDinterwordspacing

\bibitem{ma_2019}
\BIBentryALTinterwordspacing
Y.~Ma, ``Mixed-fidelity cfd simulations for aero-engines: A fan-intake interaction study,'' 2019. [Online]. Available: \url{https://www.repository.cam.ac.uk/handle/1810/297796}
\BIBentrySTDinterwordspacing

\bibitem{JIA2022107429}
\BIBentryALTinterwordspacing
Z.~Jia, H.~Tang, D.~Jin, Y.~Xiao, M.~Chen, S.~Li, and X.~Liu, ``Research on the volume-based fully coupled method of the multi-fidelity engine simulation,'' \emph{Aerospace Science and Technology}, vol. 123, p. 107429, 2022. [Online]. Available: \url{https://www.sciencedirect.com/science/article/pii/S1270963822001031}
\BIBentrySTDinterwordspacing

\bibitem{shang2019landmarks}
J.~J. Shang, ``Landmarks and new frontiers of computational fluid dynamics,'' \emph{Advances in Aerodynamics}, vol.~1, no.~1, p.~5, 2019.

\bibitem{luo2024immersed}
P.~Luo and J.~Zhang, ``An immersed boundary method coupled non-hydrostatic model for free surface flow,'' \emph{Computers and Fluids}, p. 106241, 2024.

\bibitem{RIZZI2021106940}
\BIBentryALTinterwordspacing
A.~Rizzi and J.~M. Luckring, ``Historical development and use of cfd for separated flow simulations relevant to military aircraft,'' \emph{Aerospace Science and Technology}, vol. 117, p. 106940, 2021. [Online]. Available: \url{https://www.sciencedirect.com/science/article/pii/S1270963821004508}
\BIBentrySTDinterwordspacing

\bibitem{doi:10.2514/1.A35807}
J.~G. Meeroff, D.~J. Dalle, S.~E. Rogers, A.~C. Burkhead, D.~G. Schauerhamer, and J.~F. Diaz, ``Advances in space launch system booster separation computational fluid dynamics,'' \emph{Journal of Spacecraft and Rockets}, pp. 1--13, 2024.

\bibitem{aerospace9020060}
\BIBentryALTinterwordspacing
M.~L. Erario, M.~G. De~Giorgi, and R.~Przysowa, ``Model-based dynamic performance simulation of a microturbine using flight test data,'' \emph{Aerospace}, vol.~9, no.~2, 2022. [Online]. Available: \url{https://www.mdpi.com/2226-4310/9/2/60}
\BIBentrySTDinterwordspacing

\bibitem{LECLAINCHE2023108354}
\BIBentryALTinterwordspacing
S.~{Le Clainche}, E.~Ferrer, S.~Gibson, E.~Cross, A.~Parente, and R.~Vinuesa, ``Improving aircraft performance using machine learning: A review,'' \emph{Aerospace Science and Technology}, vol. 138, p. 108354, 2023. [Online]. Available: \url{https://www.sciencedirect.com/science/article/pii/S1270963823002511}
\BIBentrySTDinterwordspacing

\bibitem{sellers1975dyngen}
J.~F. Sellers and C.~J. Daniele, \emph{DYNGEN: A program for calculating steady-state and transient performance of turbojet and turbofan engines}.\hskip 1em plus 0.5em minus 0.4em\relax National Aeronautics and Space Administration, 1975, vol. 7901.

\bibitem{veres2002overview}
J.~P. Veres, ``Overview of high-fidelity modeling activities in the numerical propulsion system simulations (npss) project,'' in \emph{Aerospace Numerical Simulation Symposium}, no. NAS 1: 15: 211351, 2002.

\bibitem{yildirim2021predicting}
F.~Yildirim~Dalkiran and M.~Toraman, ``Predicting thrust of aircraft using artificial neural networks,'' \emph{Aircraft Engineering and Aerospace Technology}, vol.~93, no.~1, pp. 35--41, 2021.

\bibitem{FAST20099}
\BIBentryALTinterwordspacing
M.~Fast, M.~Assadi, and S.~De, ``Development and multi-utility of an ann model for an industrial gas turbine,'' \emph{Applied Energy}, vol.~86, no.~1, pp. 9--17, 2009. [Online]. Available: \url{https://www.sciencedirect.com/science/article/pii/S030626190800072X}
\BIBentrySTDinterwordspacing

\bibitem{nikpey2014experimental}
H.~Nikpey, M.~Assadi, P.~Breuhaus, and P.~M{\o}rkved, ``Experimental evaluation and ann modeling of a recuperative micro gas turbine burning mixtures of natural gas and biogas,'' \emph{Applied Energy}, vol. 117, pp. 30--41, 2014.

\bibitem{li2017modelling}
W.~Li, X.~Wu, W.~Jiao, G.~Qi, and Y.~Liu, ``Modelling of dust removal in rotating packed bed using artificial neural networks (ann),'' \emph{Applied Thermal Engineering}, vol. 112, pp. 208--213, 2017.

\bibitem{KIM2020117046}
\BIBentryALTinterwordspacing
S.~Kim, K.~Kim, and C.~Son, ``Transient system simulation for an aircraft engine using a data-driven model,'' \emph{Energy}, vol. 196, p. 117046, 2020. [Online]. Available: \url{https://www.sciencedirect.com/science/article/pii/S0360544220301535}
\BIBentrySTDinterwordspacing

\bibitem{aerospace10010017}
\BIBentryALTinterwordspacing
Z.~Wang and Y.~Zhao, ``Data-driven exhaust gas temperature baseline predictions for aeroengine based on machine learning algorithms,'' \emph{Aerospace}, vol.~10, no.~1, 2023. [Online]. Available: \url{https://www.mdpi.com/2226-4310/10/1/17}
\BIBentrySTDinterwordspacing

\bibitem{karniadakis2021physics}
G.~E. Karniadakis, I.~G. Kevrekidis, L.~Lu, P.~Perdikaris, S.~Wang, and L.~Yang, ``Physics-informed machine learning,'' \emph{Nature Reviews Physics}, vol.~3, no.~6, pp. 422--440, 2021.

\bibitem{van2001art}
D.~A. Van~Dyk and X.-L. Meng, ``The art of data augmentation,'' \emph{Journal of Computational and Graphical Statistics}, vol.~10, no.~1, pp. 1--50, 2001.

\bibitem{raissi2019physics}
M.~Raissi, P.~Perdikaris, and G.~E. Karniadakis, ``Physics-informed neural networks: A deep learning framework for solving forward and inverse problems involving nonlinear partial differential equations,'' \emph{Journal of Computational physics}, vol. 378, pp. 686--707, 2019.

\bibitem{robinson2022physics}
H.~Robinson, S.~Pawar, A.~Rasheed, and O.~San, ``Physics guided neural networks for modelling of non-linear dynamics,'' \emph{Neural Networks}, vol. 154, pp. 333--345, 2022.

\bibitem{liu2022physics}
Z.~Liu, M.~Roy, D.~K. Prasad, and K.~Agarwal, ``Physics-guided loss functions improve deep learning performance in inverse scattering,'' \emph{IEEE Transactions on Computational Imaging}, vol.~8, pp. 236--245, 2022.

\bibitem{raymond2021applying}
S.~J. Raymond and D.~B. Camarillo, ``Applying physics-based loss functions to neural networks for improved generalizability in mechanics problems,'' \emph{arXiv preprint arXiv:2105.00075}, 2021.

\bibitem{bronstein2017geometric}
M.~M. Bronstein, J.~Bruna, Y.~LeCun, A.~Szlam, and P.~Vandergheynst, ``Geometric deep learning: going beyond euclidean data,'' \emph{IEEE Signal Processing Magazine}, vol.~34, no.~4, pp. 18--42, 2017.

\bibitem{cuomo2022scientific}
S.~Cuomo, V.~S. Di~Cola, F.~Giampaolo, G.~Rozza, M.~Raissi, and F.~Piccialli, ``Scientific machine learning through physics--informed neural networks: Where we are and what’s next,'' \emph{Journal of Scientific Computing}, vol.~92, no.~3, p.~88, 2022.

\bibitem{lin2023thrust}
Z.~Lin, H.~Xiao, X.~Zhang, and Z.~Wang, ``Thrust prediction of aircraft engine enabled by fusing domain knowledge and neural network model,'' \emph{Aerospace}, vol.~10, no.~6, p. 493, 2023.

\bibitem{karpatne2017physics}
A.~Karpatne, W.~Watkins, J.~Read, and V.~Kumar, ``Physics-guided neural networks (pgnn): An application in lake temperature modeling,'' \emph{arXiv preprint arXiv:1710.11431}, vol.~2, 2017.

\bibitem{cho2014properties}
K.~Cho, B.~van Merrienboer, D.~Bahdanau, and Y.~Bengio, ``On the properties of neural machine translation: Encoder--decoder approaches,'' in \emph{Proceedings of SSST-8, Eighth Workshop on Syntax, Semantics and Structure in Statistical Translation}.\hskip 1em plus 0.5em minus 0.4em\relax Association for Computational Linguistics, 2014, p. 103.

\bibitem{badrinarayanan2017segnet}
V.~Badrinarayanan, A.~Kendall, and R.~Cipolla, ``Segnet: A deep convolutional encoder-decoder architecture for image segmentation,'' \emph{IEEE Transactions on Pattern Analysis and Machine Intelligence}, vol.~39, no.~12, pp. 2481--2495, 2017.

\bibitem{chen2018encoder}
L.~C. Chen, Y.~Zhu, G.~Papandreou, F.~Schroff, and H.~Adam, ``Encoder-decoder with atrous separable convolution for semantic image segmentation,'' \emph{Computer Vision--ECCV 2018}, pp. 833--851, 2018.

\bibitem{badrinarayanan2015segnet}
V.~Badrinarayanan, A.~Handa, and R.~Cipolla, ``Segnet: A deep convolutional encoder-decoder architecture for robust semantic pixel-wise labelling,'' \emph{arXiv preprint arXiv:1505.07293}, 2015.

\bibitem{ji2021cnn}
Y.~Ji, H.~Zhang, Z.~Zhang, and M.~Liu, ``Cnn-based encoder-decoder networks for salient object detection: A comprehensive review and recent advances,'' \emph{Information Sciences}, vol. 546, pp. 835--857, 2021.

\bibitem{koh2020concept}
P.~W. Koh, T.~Nguyen, Y.~S. Tang, S.~Mussmann, E.~Pierson, B.~Kim, and P.~Liang, ``Concept bottleneck models,'' 2020.

\bibitem{vaswani2017attention}
A.~Vaswani, N.~Shazeer, N.~Parmar, J.~Uszkoreit, L.~Jones, A.~N. Gomez, {\L}.~Kaiser, and I.~Polosukhin, ``Attention is all you need,'' \emph{Advances in neural information processing systems}, vol.~30, 2017.

\bibitem{galassi2020attention}
A.~Galassi, M.~Lippi, and P.~Torroni, ``Attention in natural language processing,'' \emph{IEEE transactions on neural networks and learning systems}, vol.~32, no.~10, pp. 4291--4308, 2020.

\bibitem{han2022survey}
K.~Han, Y.~Wang, H.~Chen, X.~Chen, J.~Guo, Z.~Liu, Y.~Tang, A.~Xiao, C.~Xu, Y.~Xu \emph{et~al.}, ``A survey on vision transformer,'' \emph{IEEE transactions on pattern analysis and machine intelligence}, vol.~45, no.~1, pp. 87--110, 2022.

\bibitem{arnab2021vivit}
A.~Arnab, M.~Dehghani, G.~Heigold, C.~Sun, M.~Lu{\v{c}}i{\'c}, and C.~Schmid, ``Vivit: A video vision transformer,'' in \emph{Proceedings of the IEEE/CVF international conference on computer vision}, 2021, pp. 6836--6846.

\bibitem{10450733}
Q.~Liu, T.~Mo, Y.~Yao, and Y.~Zhang, ``Attention u-net for cell instance segmentation,'' in \emph{2023 China Automation Congress (CAC)}, 2023, pp. 1989--1994.

\bibitem{10490963}
Y.~Zhang, H.~Zhao, Z.~Yang, T.~Mo, and Y.~Yao, ``Attention-based mask r-cnn for microvascular segmentation,'' in \emph{2023 7th International Conference on Electrical, Mechanical and Computer Engineering (ICEMCE)}, 2023, pp. 961--966.

\bibitem{wang2022msran}
J.~Wang, L.~Yu, and S.~Tian, ``Msran: A multi-scale residual attention network for multi-model image fusion,'' \emph{Medical \& Biological Engineering \& Computing}, vol.~60, no.~12, pp. 3615--3634, 2022.

\bibitem{wang2022amfnet}
J.~Wang, L.~Yu, S.~Tian, W.~Wu, and D.~Zhang, ``Amfnet: An attention-guided generative adversarial network for multi-model image fusion,'' \emph{Biomedical Signal Processing and Control}, vol.~78, p. 103990, 2022.

\bibitem{derose2020attention}
J.~F. DeRose, J.~Wang, and M.~Berger, ``Attention flows: Analyzing and comparing attention mechanisms in language models,'' \emph{IEEE Transactions on Visualization and Computer Graphics}, vol.~27, no.~2, pp. 1160--1170, 2020.

\bibitem{xu2021adaptive}
T.~Xu, Z.~Feng, X.-J. Wu, and J.~Kittler, ``Adaptive channel selection for robust visual object tracking with discriminative correlation filters,'' \emph{International Journal of Computer Vision}, vol. 129, pp. 1359--1375, 2021.

\bibitem{gao2019change}
Y.~Gao, F.~Gao, J.~Dong, and S.~Wang, ``Change detection from synthetic aperture radar images based on channel weighting-based deep cascade network,'' \emph{IEEE journal of selected topics in applied earth observations and remote sensing}, vol.~12, no.~11, pp. 4517--4529, 2019.

\bibitem{hu2018squeeze}
J.~Hu, L.~Shen, and G.~Sun, ``Squeeze-and-excitation networks,'' in \emph{Proceedings of the IEEE conference on computer vision and pattern recognition}, 2018, pp. 7132--7141.

\bibitem{weng2022deep}
X.~Weng, Y.~Yan, G.~Dong, C.~Shu, B.~Wang, H.~Wang, and J.~Zhang, ``Deep multi-branch aggregation network for real-time semantic segmentation in street scenes,'' \emph{IEEE Transactions on Intelligent Transportation Systems}, vol.~23, no.~10, pp. 17\,224--17\,240, 2022.

\bibitem{he2020multi}
Y.~He, L.~Dai, and H.~Zhang, ``Multi-branch deep residual learning for clustering and beamforming in user-centric network,'' \emph{IEEE communications letters}, vol.~24, no.~10, pp. 2221--2225, 2020.

\bibitem{Kingma2014AdamAM}
\BIBentryALTinterwordspacing
D.~P. Kingma and J.~Ba, ``Adam: A method for stochastic optimization,'' \emph{CoRR}, vol. abs/1412.6980, 2014. [Online]. Available: \url{https://api.semanticscholar.org/CorpusID:6628106}
\BIBentrySTDinterwordspacing

\end{thebibliography}
\bibliographystyle{IEEEtran}

\begin{IEEEbiography}[{\includegraphics[width=1in,height=1.25in,clip,keepaspectratio]{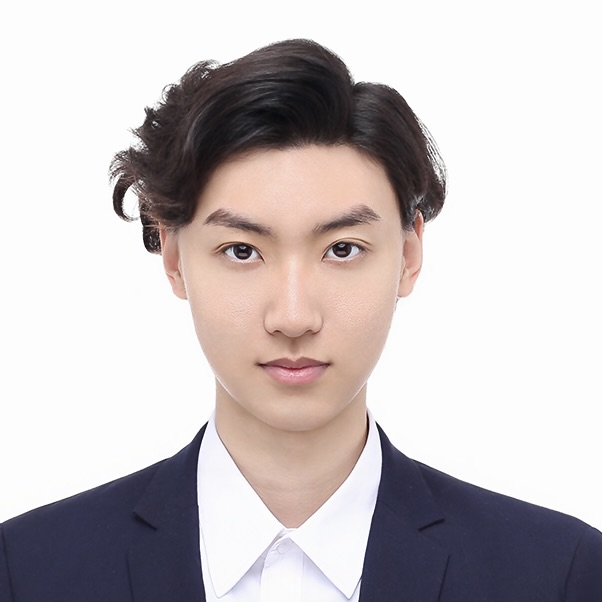}}]{Tong Mo} received the B.S. degree in internet of things engineering from the Beijing University of Technology, Beijing, China, in 2024. He is currently working toward M.S. degree in computer engineering in the New York University, USA.

From 2023 to 2024, he was a Research Assistant for the State Key Laboratory of Multimodal Artificial Intelligence Systems, Institute of Automation, Chinese Academy of Sciences, Beijing, China. His research interest include Industrial applied machine learning, few-shot learning and embedded system.

\end{IEEEbiography}%

\begin{IEEEbiography}[{\includegraphics[width=1in,height=1.25in,clip,keepaspectratio]{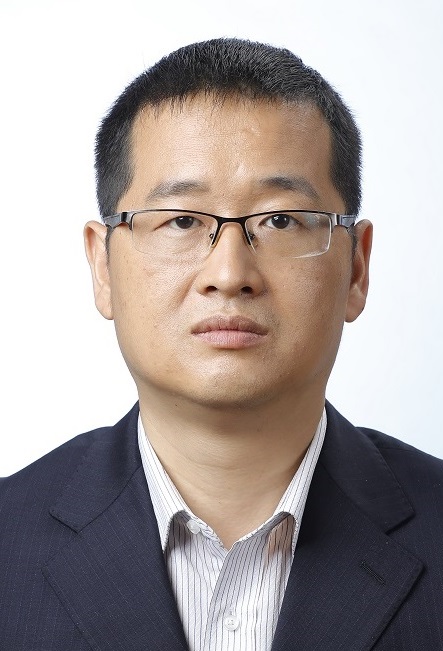}}]{Shuxiao Li} received the B.S. degree in mechanical engineering and automation from Xi’an Jiaotong University, Xi’an, China, in 2003, and the Ph.D. degree in pattern recognition and intelligent systems from the University of Chinese Academy of Sciences, Beijing, China, in 2008.

He is currently an associate researcher at the State Key Laboratory of Multimodal Artificial Intelligence Systems, Institute of Automation, Chinese Academy of Sciences, Beijing, China. He is also a master's supervisor at the School of Artificial Intelligence, University of Chinese Academy of Sciences, Beijing, China. He has authored over 50 journal articles and conference papers. His main research interests include machine learning, collaborative perception, and vision systems.
\end{IEEEbiography}

\begin{IEEEbiography}[{\includegraphics[width=1in,height=1.25in,clip,keepaspectratio]{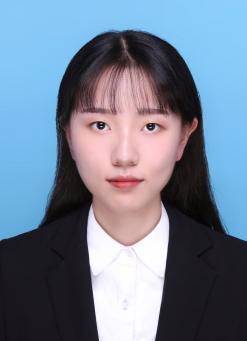}}]{Xiaomeng Zhu} received the B.S. degree in automation from the University of Electronic Science and Technology of China in 2021. She obtained her M.S. degree in pattern recognition and intelligent systems from the University of Chinese Academy of Sciences in 2024. 

Her research interests include few-shot learning, incremental learning and embodied artificial intelligence.
\end{IEEEbiography}
\end{document}